# Matrix Coherence and the Nyström Method

**Ameet Talwalkar**
Courant Institute of Mathematical Sciences
New York University
ameet@cs.nyu.edu

**Afshin Rostamizadeh**
Courant Institute of Mathematical Sciences
New York University
rostami@cs.nyu.edu

## Abstract

The Nyström method is an efficient technique used to speed up large-scale learning applications by generating low-rank approximations. Crucial to the performance of this technique is the assumption that a matrix can be well approximated by working exclusively with a subset of its columns. In this work we relate this assumption to the concept of matrix coherence, connecting coherence to the performance of the Nyström method. Making use of related work in the compressed sensing and the matrix completion literature, we derive novel coherence-based bounds for the Nyström method in the low-rank setting. We then present empirical results that corroborate these theoretical bounds. Finally, we present more general empirical results for the full-rank setting that convincingly demonstrate the ability of matrix coherence to measure the degree to which information can be extracted from a subset of columns.

## 1 Introduction

Modern problems in computer vision, natural language processing, computational biology and other areas often involve datasets containing millions of training instances. However, several standard methods in machine learning, such as spectral clustering (Ng *et al.*, 2001), manifold learning techniques (de Silva and Tenenbaum, 2003; Schölkopf *et al.*, 1998), kernel ridge regression (Saunders *et al.*, 1998) or other kernel-based algorithms do not scale to such orders of magnitude. In fact, even storage of the matrices associated with these datasets can be problematic since they are often not sparse and hence the number of entries is extremely large. As shown by Williams and Seeger (2000), the Nyström method provides an attractive solution when working with large-scale datasets by operating on only a small part of the original matrix to generate a low-rank approximation. The Nyström method has been shown to work well in practice for various applications ranging from manifold learning to image segmentation (Fowlkes *et al.*, 2004; Platt, 2004; Talwalkar *et al.*, 2008; Zhang *et al.*, 2008).

The effectiveness of the Nyström method hinges on two key assumptions on the input matrix, $G$. First, we assume that a low-rank approximation to $G$ can be effective for the task at hand. This assumption is often true empirically as evidenced by the widespread use of singular value decomposition (SVD) and principal component analysis (PCA) in practical applications. As expected, the Nyström method is not appropriate in cases where this assumption does not hold, which explains its poor performance in the experimental results of Fergus *et al.* (2009). Previous work analyzing the performance of the Nyström method incorporates this low-rank assumption into theoretical guarantees by comparing the Nyström approximation to the 'best' low-rank approximation, i.e., the approximation constructed from the top singular values and singular vectors of $G$ (see Section 2 for further discussion) (Drineas and Mahoney, 2005; Kumar *et al.*, 2009a).

The second crucial assumption of the Nyström method involves the sampling-based nature of the algorithm, namely that an accurate low-rank approximation can be generated exclusively from information extracted from a small subset of $l \ll n$ columns of $G$. This assumption is not generally true for all matrices. For instance, consider the extreme case of the $n \times n$ matrix described below:

$$G = \begin{bmatrix} | & & | & | & & | \\ \vec{e}_1 & \dots & \vec{e}_r & \vec{0} & \dots & \vec{0} \\ | & & | & | & & | \end{bmatrix}, \tag{1}$$

where $\vec{e}_i$ is the $i$th column of the $n$ dimensional identity matrix and $\vec{0}$ is the $n$ dimensional zero vector.

Although this matrix has rank $r$, it nonetheless cannot be well approximated by a random subset of $l$ columns unless this subset includes $e_1, \ldots, e_r$. In order to account for such pathological cases, previous theoretical bounds relied on sampling columns of $G$ from a non-uniform distribution weighted precisely by the magnitude of the diagonal elements of $G$ (Belabbas and Wolfe, 2009; Drineas and Mahoney, 2005). Indeed, these bounds give better guarantees for pathological cases. However, in practice, when working with real-world datasets, uniform sampling is more commonly used, e.g., Fowlkes *et al.* (2004); Platt (2004); Talwalkar *et al.* (2008); Williams and Seeger (2000), since diagonal sampling is more expensive and does not typically outperform uniform sampling (Kumar *et al.*, 2009c). Hence the diagonal sampling bounds are not applicable in this setting. Furthermore, these bounds are typically loose for matrices in which the diagonal entries of the matrix are roughly of the same magnitude, as in the case of all kernel matrices generated from RBF kernels, for which the Nyström has been noted to work particularly well (Williams and Seeger, 2000).

In this work, we propose to characterize the ability to extract information from a small subset of $l$ columns using the notion of matrix *coherence*, an alternative data-dependent measurement which we believe to be intrinsically related to the algorithm's performance. Coherence measures the extent to which the singular vectors of a matrix are correlated with the standard basis. Intuitively, if we work with sufficiently incoherent matrices, then we avoid pathological cases such as the one presented (1). Recent work on compressed sensing and matrix completion, which also involve sampling-based approximations, have relied heavily on coherence assumptions (Candès and Romberg, 2007; Candès *et al.*, 2006; Donoho, 2006).

The main contribution of this work is the connection that is made between matrix coherence and the Nyström method. Making use of related work in the compressed sensing and the matrix completion literature, we give a more refined analysis of this algorithm as a function of matrix coherence, presenting a novel preliminary theoretical bound for the Nyström method. We also present extensive empirical results that strongly relate coherence to the performance of the Nyström method.

The remainder of the paper is organized as follows. Section 2 introduces basic definitions of coherence and gives a brief presentation of the Nyström method. In Section 3 we present our novel bound for the Nyström method under low-rank, low-coherence assumptions. Section 4 presents extensive empirical studies that support our bound and illustrate a similar connection between matrix coherence and the performance of the Nyström method for full-rank matrices. Our empirical results also show that incoherence assumptions are valid for several datasets derived from real-world applications.

## 2 Preliminaries

Let $G \in \mathbb{R}^{n\times n}$ be a symmetric positive semidefinite (SPSD) matrix. SPSD matrices, such as Gram or kernel matrices, often appear in the context of machine learning. For any Gram matrix, there exists an $N$ and $X \in \mathbb{R}^{N\times n}$ such that $G = X^\top X$. We define $X^{(j)}$, $j = 1 \ldots n$, as the $j$th column vector of $X$ and $X_{(i)}$, $i = 1 \ldots N$, as the $i$th row vector of $X$, and denote by $\|\cdot\|$ the $\ell_2$ norm of a vector. Using singular value decomposition (SVD), the Gram matrix can be written as $G = V\Sigma V^\top$, where $V$ is orthonormal and $\Sigma = \mathrm{diag}(\sigma_1, \ldots, \sigma_n)$ is a real diagonal matrix with diagonal entries sorted in decreasing order. For $r = \mathrm{rank}(G)$, the pseudo-inverse of $G$ is defined as $G^+ = \sum_{t=1}^{r} \sigma_t^{-1} V^{(t)} {V^{(t)}}^\top$. Further, for $k \le r$, $G_k = \sum_{t=1}^{k} \sigma_t V^{(t)} {V^{(t)}}^\top$ is the 'best' rank-$k$ approximation to $G$, or the rank-$k$ matrix with minimal $\|\cdot\|_\xi$ distance to $G$, where $\xi \in \{2, F\}$ and $\|\cdot\|_2$ denotes the spectral norm and $\|\cdot\|_F$ the Frobenius norm of a matrix.

### 2.1 Nyström method

The Nyström method was presented in Williams and Seeger (2000) to speed up the performance of kernel machines. This is done by generating low-rank approximations of $G$ using a subset of the columns of the matrix. Suppose we randomly sample $l \ll n$ columns of $G$ uniformly with replacement, and let $C$ be the $n \times l$ matrix of these sampled columns. Then, without loss of generality, we can rearrange the columns and rows of $G$ based on this sampling and define $X = [X_1 \quad X_2]$ where $X_1 \in \mathbb{R}^{N\times l}$, such that

$$G = X^\top X = \begin{bmatrix} W & X_1^\top X_2 \\ X_2^\top X_1 & X_2^\top X_2 \end{bmatrix} \tag{2}$$

$$\text{and} \quad C = \begin{bmatrix} W \\ X_2^\top X_1 \end{bmatrix}, \tag{3}$$

where $W = X_1^\top X_1$. The Nyström approximation is now defined as:

$$G \approx \widetilde{G} = C W^+ C^\top. \tag{4}$$

The Frobenius distance between $G$ and $\widetilde{G}$, $\|G - \widetilde{G}\|_F$, is one standard measurement of the accuracy of the Nyström method. The runtime of this algorithm is $O(l^3 + nl^2)$: $O(l^3)$ for SVD on $W$ and $O(nl^2)$ for multiplication with $C$. The Nyström method is often presented with an additional step whereby $W$ in (4) is replaced by its rank-$k$ approximation, $W_k$, for some $k < l$, thus generating $\widetilde{G}_k$, the rank-$k$ Nyström approximation to $G$. In this case, the runtime of the algorithm is reduced to $O(l^3 + nlk)$.

### 2.2 Coherence

Although the Nyström method tends to work well in practice, the performance of this algorithm depends on the structure of the underlying matrix. We will show that the performance is related to the size of the entries of the singular vectors of $G$, or the *coherence* of its singular vectors. We define $V_r$ as the top $r$ singular vectors of $G$, and denote the coherence of these singular vectors as $\mu(V_r)$, which is adapted from Candès and Romberg (2007).

**Definition 1** (Coherence)**.** *The* coherence *of a matrix of* $V_r$ *with orthonormal columns is defined as:*

$$\mu(V_r) = \sqrt{n} \max_{i,j} |V_{r(i)}^{(j)}| \, . \tag{5}$$

The coherence of $V_r$ is lower bounded by 1, as is the case for the rank-1 matrix with all entries equal to $1/\sqrt{n}$, and upper bounded by $\sqrt{n}$, as is the case for the matrix of canonical basis vectors. As discussed in Candès and Recht (2009); Candès and Tao (2009), highly coherent matrices are difficult to randomly recover via matrix completion algorithms, and this same logic extends to the Nyström method. In contrast, incoherent matrices are much easier to successfully complete and to approximate via the Nyström method, as discussed in Section 3.

In order to provide some intuition, Candès and Recht (2009) give several classes of randomly generated matrices with low coherence. One such class of matrices is generated from uniform random orthonormal singular vectors and arbitrary singular values. For such a class they show that $\mu = O(\sqrt{\log n} \cdot \sqrt[4]{r})$ with high probability.[1] In what follows, we will show bounds on the number of points needed for reconstruction that become more favorable as coherence decreases. However, the bounds are useful for more generous values of coherence than given in the above example. We will also provide an empirical study of coherence for various real-world and synthetic examples.

[1] For low-rank matrices, $\sqrt[4]{r}$ is quite small. Moreover, this $\sqrt[4]{r}$ factor only appears due to our use of the generally loose inequality $\mu^2 \leq \sqrt{r}\mu_1$, where $\mu_1$ is a slightly different notion of coherence used in the original bound in Candès and Recht (2009) for this class of matrices.

## 3 Low-rank, low-coherence bounds

In this section, we make use of coherence to analyze the Nyström method when used with low-rank matrices. We note that although the bounds presented throughout this section hold for matrices of any rank $r$, they are only interesting when $r = o(\sqrt{n})$, and hence they are most applicable in the "low-rank" setting.

### 3.1 Nyström method bound

The Nyström method is empirically effective in cases where $G$ has low-rank structure even if the matrix has full-rank, i.e., $G \approx G_k$ for some $k \ll n$. Furthermore, as stated in Theorem 1 below, when $G$ is actually a low-rank matrix, then the Nyström method can exactly recover the initial matrix (we include the short proof for the sake of completeness).

**Theorem 1** ((Kumar *et al.*, 2009b) Thm. 3)**.** *Suppose* $r = \text{rank}(G) \leq k \leq l$ *and* $\text{rank}(W) = r$*. Then the Nyström approximation is exact, i.e.,* $\|G - \widetilde{G}_k\|_F = 0$.

*Proof.* Since $G = X^\top X$, $\text{rank}(G) = \text{rank}(X) = r$. Similarly, $W = X_1^\top X_1$ implies $\text{rank}(X_1) = r$, i.e., the columns of $X_1$ span the columns of $X$. We next let $U_{X_1,k}$ be the $k$ left singular vectors of $X_1$ associated with the top $k$ singular values of $X_1$. We then represent $W$ and $C$ in terms of $X_1$ and $X_2$, to rewrite the Nyström approximation as:

$$\begin{aligned} \widetilde{G} &= C W_k^+ C^\top \\ &= \begin{bmatrix} X_1^\top \\ X_2^\top \end{bmatrix} X_1 (X_1^\top X_1)_k^+ X_1^\top \begin{bmatrix} X_1 & X_2 \end{bmatrix} \\ &= X^\top U_{X_1,k} U_{X_1,k}^\top X. \end{aligned} \tag{6}$$

Furthermore, since columns of $X_1$ span the columns of $X$, $U_{X_1,r}$ is an orthonormal basis for $X$ and $I - U_{X_1,r}U_{X_1,r}^\top$ is an orthogonal projection matrix into the nullspace of $X$. Since $k \geq r$, from (6) we have

$$\|G - \widetilde{G}_k\|_F = \|X^\top (I - U_{X_1,k} U_{X_1,k}^\top) X\|_F = 0. \tag{7}$$

□

This theorem implies that if $G$ has low-rank, then *there exists* a particular sampling such that $\text{rank}(W) = \text{rank}(G)$ and the Nyström method can perfectly recover the full matrix. However, selecting a suitable set of $l$ columns from an $n \times n$ SPSD matrix can be an intractable combinatorial problem, and there exist matrices for which the probability of selecting such a subset uniformly at random is exponentially small, e.g., the rank-$r$ SPSD diagonal matrices discussed earlier. In contrast, a large class of SPSD matrices are much more incoherent, and for these matrices, we will next show that by choosing $l$ to be linear in $r$ and logarithmic in $n$ we can can with very high probability guarantee that $\text{rank}(W) = r$, and hence exactly recover the initial matrix.

### Probability of choosing a good subset

We start with a rank-$r$ Gram matrix, $G$, and a fixed distribution, $\mathcal{D}$, over the columns of $G$. Our goal is to calculate the probability of randomly choosing a subset of $l$ columns of $G$ according to $\mathcal{D}$ such that $\text{rank}(W) = r$. Recall that $G = X^\top X$, $X = [X_1 \quad X_2]$ and $W = X_1^\top X_1$. Then, by properties of SVD, we know that $\text{rank}(G) = \text{rank}(X)$ and $\text{rank}(W) = \text{rank}(X_1)$. Hence, the probability of this desired event is equivalent to the probability of sampling $l$ columns of $X$ according to $\mathcal{D}$ such that $\text{rank}(X_1) = r$, as shown in (10). Next, we can write the thin SVD of $X$ as $X = U_r \Sigma_r V_r^\top$, where $U_r \in \mathbb{R}^{m \times r}$, $\Sigma_r \in \mathbb{R}^{r \times r}$ and $V_r \in \mathbb{R}^{n \times r}$. Since $U_r$ contains orthonormal columns and $\Sigma_r$ is invertible, we know that

$$\Sigma_r^{-1} U_r^\top X = V_r^\top . \tag{8}$$

Further, using the block representation of $X$, we have

$$X_1^\top U_r \Sigma_r^{-1} = V_{r,l}, \tag{9}$$

where $V_{r,l} \in \mathbb{R}^{l \times r}$ corresponds to the first $l$ components for each of the $r$ right singular vectors of $X$. Since $\text{rank}(X_1) = \text{rank}(X_1^\top U_r \Sigma_r^{-1})$, we obtain the equality of (11).

$$\begin{aligned} \Pr_{\mathcal{D}}[\text{rank}(W) = r] &= \Pr_{\mathcal{D}}[\text{rank}(X_1) = r] & (10) \\ &= \Pr_{\mathcal{D}}[\text{rank}(V_{r,l}) = r]. & (11) \end{aligned}$$

In the next section we calculate this probability for a specific distribution in terms of $l$ as well as a measure of the coherence of $V_r$.

### Sampling Bound

Given the orthonormal matrix $V_r$, we would like to find a choice of $l$ such that $V_{r,l}$ created by *uniform* sampling has rank $r$ with high probability. As pointed out in the previous section, a meaningful bound may not be possible for any $l < n$ if no assumption is made on $V_r$. Here we adopt the assumption that $V_r$ has low coherence, as defined in Definition 1. We then observe that by properties of SVD we have

$$\Pr\Big(\text{rank}(V_{r,l}) = r\Big) = \Pr\Big(\text{rank}(V_{r,l}^\top V_{r,l}) = r\Big) . \tag{12}$$

Next, we define $\sigma = \|V_{r,l}^\top V_{r,l}\|_2$ and note that for $0 < c < 1/\sigma$, $cV_{r,l}^\top V_{r,l}$ is an $r \times r$ SPSD matrix with singular values less than one. Furthermore, $I - cV_{r,l}^\top V_{r,l}$ is also SPSD with

$$\Pr\Big(\text{rank}(V_{r,l}^\top V_{r,l}) = r\Big) = \Pr\Big(\|cV_{r,l}^\top V_{r,l} - I\| < 1\Big) , \tag{13}$$

since $\|cV_{r,l}^\top V_{r,l} - I\| = 1$ implies that the nullspace of $cV_{r,l}^\top V_{r,l}$ is nonempty. Alternatively, if $c \geq 1/\sigma$, then

$$\Pr\Big(\text{rank}(V_{r,l}^\top V_{r,l}) = r\Big) \geq \Pr\Big(\|cV_{r,l}^\top V_{r,l} - I\| < 1\Big) , \tag{14}$$

since, for large enough $c$, we could have $\|cV_{r,l}^\top V_{r,l} - I\| \geq 1$ even if $\text{rank}(V_{r,l}^\top V_{r,l}) = r$. Thus the inequality in (14) holds for any constant $c > 0$, i.e., the probability on the RHS of (14) serves as a lower bound for the probability of interest to us.

The probability on the RHS of (14) has been studied in previous compressive sampling literature. Specifically, Candès and Romberg (2007) makes use of a main lemma of Rudelson (1999) to derive Theorem 2, which provides us with our desired lower bound.

**Theorem 2** ( (Candès and Romberg, 2007) Thm. 1.2)**.** *Define $V_r \in \mathbb{R}^{n \times r}$ such that $V_r^\top V_r = I$ and let $V_{r,l} \in \mathbb{R}^{l \times r}$ be generated from $V_r$ by sampling rows uniformly at random. Then, the following holds with probability at least $1 - \delta$,*

$$\Big\| \frac{n}{l} V_{r,l}^\top V_{r,l} - I \Big\| < \frac{1}{2} , \tag{15}$$

*for any $l$ that satisfies,*

$$l \geq r\mu^2(V_r) \max\big(C_1 \log(r), C_2 \log(3/\delta)\big), \tag{16}$$

*where $C_1$ and $C_2$ are positive constants.*

Note that our definition of coherence and statement of Theorem 2 are modified to account for the fact that $V_r^\top V_r = I$ as oppose to $nI$, as in Candès and Romberg (2007). Also, $V_r$ is not square as assumed in the original theorem, however it can be verified that the proof holds even for this case.

By making use of Theorem 2, we can now answer the question regarding the number of columns needed to sample from $G$ in order to obtain an exact reconstruction via the Nyström method. Theorem 3 presents a bound on $l$ for matrix completion in terms of $\mu$.

**Theorem 3.** *Let $G \in \mathbb{R}^{n \times n}$ be a rank-$r$ SPSD matrix where $V_r \in \mathbb{R}^{n \times r}$ is a matrix of its singular vectors. Define $\widetilde{G}_k$ as the Nyström approximation of $G$ using $l$ randomly sampled columns of $G$, with $l \geq k \geq r$. Then it suffices to sample $l \geq r\mu^2(V_r) \max\big(C_1 \log(r), C_2 \log(3/\delta)\big)$ columns, where $C_1$ and $C_2$ are positive constants, to have with probability at least $1 - \delta$,*

$$\|G - \widetilde{G}_k\| = 0 . \tag{17}$$

| Dataset | Type of data | # Points ($n$) | # Features ($d$) | Kernel |
|---|---|---|---|---|
| PIE (Sim *et al.*, 2002) | face images | 2731 | 2304 | linear |
| MNIST (LeCun and Cortes, 1998) | digit images | 4000 | 784 | linear |
| Essential (Gustafson *et al.*, 2006) | proteins | 4728 | 16 | RBF |
| Abalone (Asuncion and Newman, 2007) | abalones | 4177 | 8 | RBF |
| Dexter (Asuncion and Newman, 2007) | bag of words | 2000 | 20000 | linear |
| Random | random features | 1000 | 20000 | linear |
| Artificial | high coherence | 2000 | - | - |

Table 1: A summary of the datasets used in the experiments, including the type of data, the number of points ($n$), the number of features ($d$) and the choice of kernel.

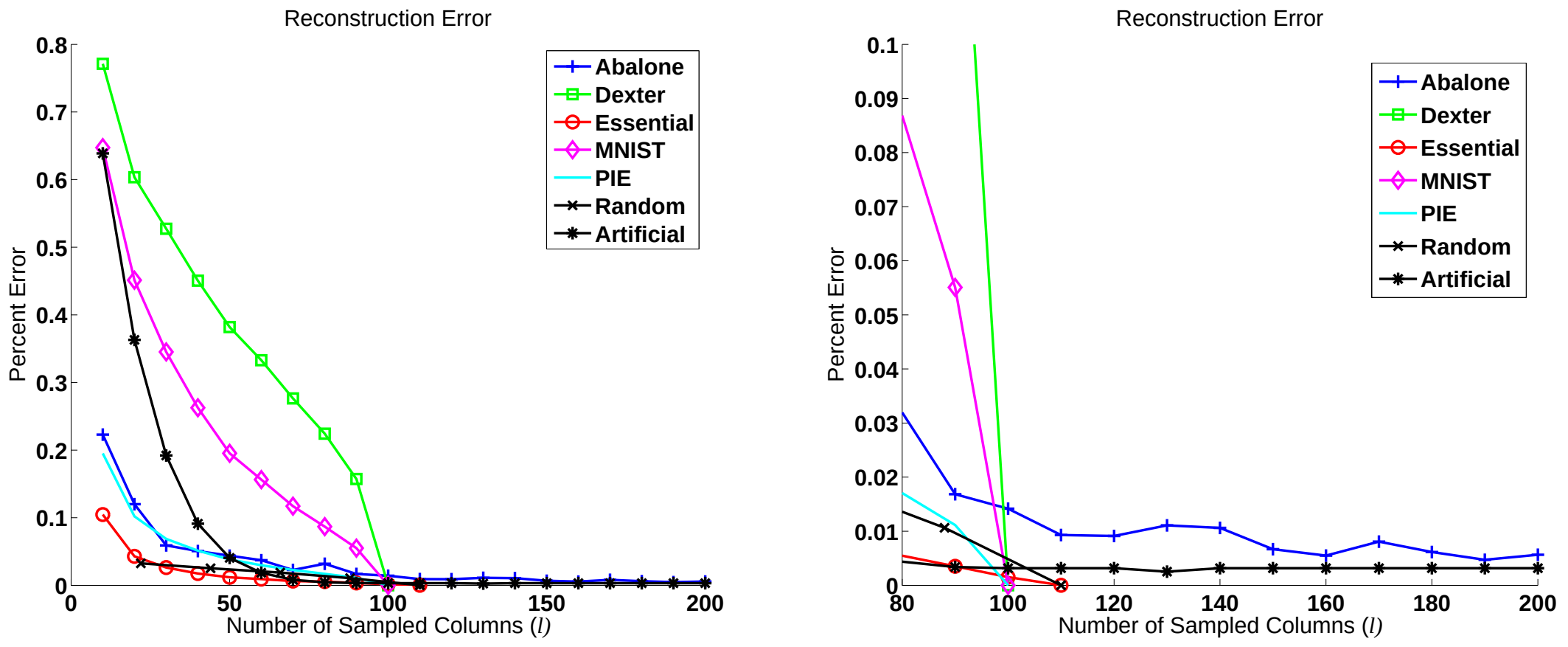

Figure 1: Mean percent error over 10 trials of Nyström approximations of rank 100 matrices. Left: Results for $l$ ranging from 5 to 200. Right: Magnified view of experimental results for $l$ ranging from 80 to 200.

*Proof.* Theorem 1 states sufficient conditions for exact matrix completion. Equations (10) and (11) reduce these sufficient conditions to a condition on the rank of $V_{r,l}$. Equations (12) and (14) further reduce this problem to a similar problem previously studied in the context of compressed sensing. Finally, we use Theorem 2 to bound with high probability the RHS of (14). □

## 4 Experiments

In this section we present a series of empirical results that show the empirical connection between matrix coherence and the performance of the Nyström method. We first perform two sets of experiments that corroborate the theoretical claims made in the previous section – Section 4.1 illustrates the performance of the Nyström method for low-rank matrices using the seven datasets detailed in Table 1 while Section 4.2 interprets these results in the context of the coherence of these datasets. Next, we present more general experimental results in Section 4.3 that connect matrix coherence to the Nyström method in the case of full-rank matrices.

### 4.1 Reconstruction error

In our first set of experiments we measure the accuracy of the Nyström approximation ($\widetilde{G}_k$) for a variety of rank-$r$ matrices, with $r = 100$. For the first six datasets listed above, we initially constructed the optimal rank-$r$ approximation to each kernel matrix by reconstructing with the top $r$ eigenvalues and eigenvectors. The final rank-$r$ SPSD matrix (Artificial) was constructed to have high coherence, i.e., large $\mu$, following the procedure outlined in Section 4.3.[2] Next, we performed the Nyström method for various values of $l$ to generate a series of approximations to our rank-$r$ matrices (note that we set $k = l$). For each approximation, we calculated the percent error of the Nyström approximation using the notion of percent error, defined as follows:

$$\text{Percent error} = \frac{\|G - \widetilde{G}_k\|_F}{\|G\|_F} \times 100. \qquad (18)$$

The results of this experiment, averaged over 10 trials, are presented in Figure 1. The figure shows that for five of the seven datasets, the Nyström method exactly reconstructs the initial rank $r$ matrix when the number of sampled columns ($l$) is equal or slightly larger than $r$. Note that this observation holds for each of the ten trials, since the mean error is zero for each of these datasets when $l \approx r$. In contrast, for the cases of the Abalone and Artificial datasets, we do not see convergence to zero percent error as $l$ surpasses $r$, and the percent error is non-zero even when $l = 2r$.

[2]We used an artificial dataset due to the difficulty of finding multiple real examples with large $\mu$.

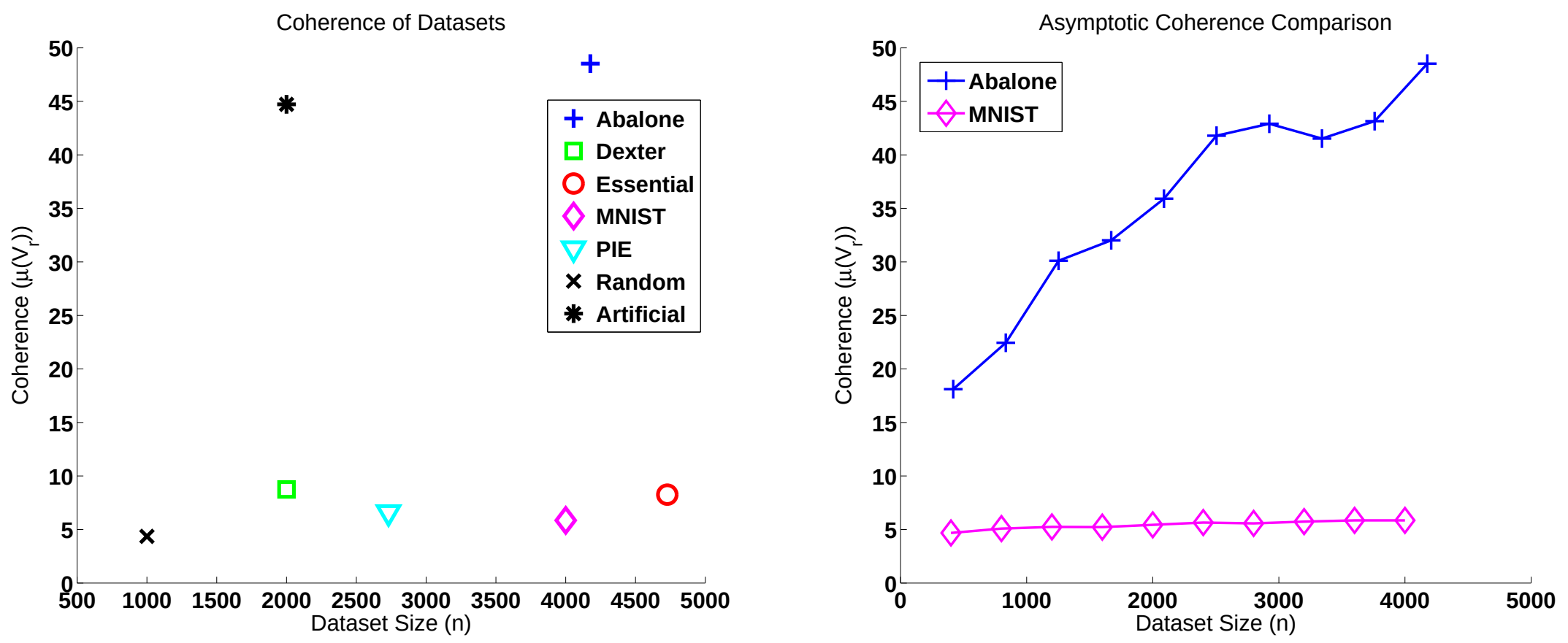

Figure 2: Coherence of Datasets. Left: Coherence of rank 100 SPSD matrices used in experiments in Section 4.1. Right: Asymptotic growth of coherence for MNIST and Abalone datasets. Note that coherence values are means over ten trials.

### 4.2 Coherence of datasets

In this set of experiments, we use the concept of coherence to explain the results from Section 4.1, namely that the Nyström method generates an exact matrix reconstruction for $l \approx r$ for five of the seven datasets, but fails to do so for the Abalone and Artificial datasets. As such, we first calculated the coherence of each of the SPSD rank 100 matrices used in Section 4.1, using the definition of coherence from Definition 1. The left panel of Figure 2 shows the coherence of these matrices with respect to the number of points in the dataset. This plot illustrates the stark contrast between Abalone/Artificial and the other five datasets in terms of coherence, and helps validate our theoretical connection between low-coherence matrices and the ability to generate exact reconstructions via the Nyström method.

Next, we performed an experiment in which we repeatedly subsampled the initial SPSD matrices to generate matrices with different dimensions, i.e., different values of $n$. For each value of $n$, we computed the coherence of the subsampled matrix, again using Definition 1. The right panel of Figure 2 shows the mean results over ten trials for both the MNIST and Abalone datasets. As illustrated by this plot, the coherence of the Abalone dataset grows much more quickly than that of the MNIST dataset. As illustrated by the orthogonal random model, we expect incoherent matrices to exhibit a slow rate of growth, i.e. $O(\sqrt{\log n} \cdot \sqrt[4]{r})$. The plots for the Artificial dataset are comparable to the Abalone dataset, and the plots for the other four datasets are comparable to the MNIST dataset (plots not shown). These results provide further intuition for why the Nyström method is able to perform exact reconstruction on all datasets except for Abalone.

### 4.3 Full rank experiments

As discussed in Section 1, the Nyström method hinges on two assumptions: good low-rank structure of the matrix and the ability to extract information from a small subset of $l$ columns of the input matrix. In this section, we analyze the effect of each of these assumptions on Nyström method performance on full-rank matrices, using matrix coherence as a quantification of the latter assumption. To do so, we devised a series of experiments using synthetic datasets that precisely control the effects of each of these parameters.

To control the low-rank structure of the matrix, we generated artificial datasets with exponentially decaying eigenvalues with differing decay rates, i.e., for $i \in \{1, \ldots, n\}$ we defined the $i$th singular value as $\sigma_i = \exp(-i\eta)$, where $\eta$ controls the rate of decay. For a fixed value of $\eta$, we then measured the percentage of the spectrum captured by the top $k$ singular values as follows:

$$\text{Percent of Spectrum} = \frac{\sum_{i=1}^{k} \sigma_i}{\sum_{i=1}^{n} \sigma_i}. \qquad (19)$$

To control coherence, we generated singular vectors with varying coherences by forcing the first singular vector to achieve our desired coherence and then using QR to generate a full orthogonal basis. The smallest values of $\mu$ used in our experiments correspond to randomly generated orthogonal matrices. We report the results of our experiments in Figure 3. For these experiments we set $n = 2000$ and $k = 50$. Each plot corresponds to matrices with a fixed eigenvalue decay rate (resulting in a fixed percentage of spectrum captured) and each line within a plot corresponds to the average results of 10 randomly generated matrices with the specified coherence. Furthermore, results for each such matrix for a fixed percentage of sampled columns are the means over 5 random subsets of columns.

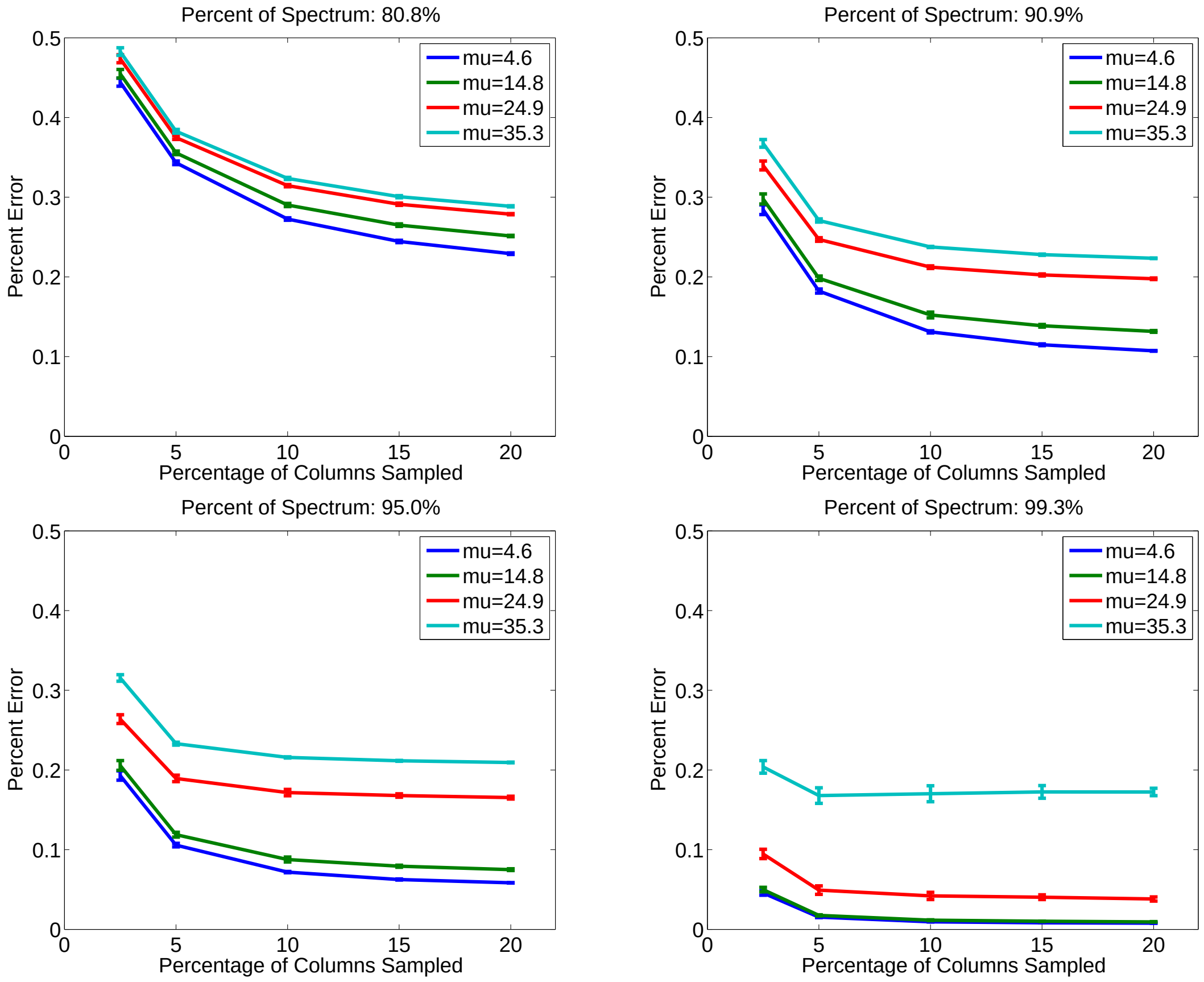

Figure 3: Coherence experiments with full-rank synthetic datasets, with $n = 2000$ and $k = 50$. Each plot corresponds to matrices with a fixed eigenvalue decay rate (resulting in a fixed percentage of spectrum captured) and each line within a plot corresponds to the average results of 10 randomly generated matrices with the specified coherence. Furthermore, results for each such matrix for a fixed percentage of sampled columns are the means over 5 random subsets of columns.

There are two main observations to be drawn from our experiments. First, as noted in previous work with the Nyström method, the Nyström method generates better approximations for matrices with better low rank structure, i.e., matrices with a higher percentage of spectrum captured by the top $k$ singular values. Second, following the same pattern as in the low-rank setting, the Nyström method generates better approximations for lower coherence matrices, and hence, matrix coherence appears to effectively capture the degree to which information can be extracted from a subset of columns.

## 5 Conclusion and future work

In this work, we make a connection between matrix coherence and the performance of the Nyström method. Making use of related work in the compressed sensing and the matrix completion literature, we derive novel coherence-based bounds for the Nyström method in the low-rank setting. We then present empirical results that corroborate these theoretical bounds. Finally, we present more general empirical results for the full-rank setting that convincingly demonstrate the ability of matrix coherence to measure the degree to which information can be extracted from a subset of columns.

Future work involves generalizing our coherence-based bounds to the case of full-rank matrices. Additionally, our preliminary studies suggest that matrix coherence can perhaps be estimated using a subset of matrix entries. Developing an algorithm to efficiently estimate matrix coherence is of crucial importance to help quickly determine the applicability of the Nyström method (and other algorithms with performance tied to matrix coherence) on a case-by-case basis.

## References

A. Asuncion and D.J. Newman. UCI machine learning repository. `http://www.ics.uci.edu/~mlearn/MLRepository.html`, 2007.

M. A. Belabbas and P. J. Wolfe. Spectral methods in machine learning and new strategies for very large datasets. *Proceedings of the National Academy of Sciences of the United States of America*, 106(2):369–374, January 2009.

Emmanuel J. Candès and Benjamin Recht. Exact matrix completion via convex optimization. *Foundations of Computational Mathematics*, 9(6):717–772, 2009.

E. J. Candès and J. Romberg. Sparsity and incoherence in compressive sampling. *Inverse Problems*, 23(3):969–986, 2007.

Emmanuel J. Candès and Terence Tao. The power of convex relaxation: Near-optimal matrix completion. `arXiv:0903.1476v1[cs.IT]`, 2009.

Emmanuel J. Candès, Justin K. Romberg, and Terence Tao. Robust uncertainty principles: exact signal reconstruction from highly incomplete frequency information. *IEEE Transactions on Information Theory*, 52(2):489–509, 2006.

Vin de Silva and Joshua Tenenbaum. Global versus local methods in nonlinear dimensionality reduction. In *Neural Information Processing Systems*, 2003.

David L. Donoho. Compressed Sensing. *IEEE Transactions on Information Theory*, 52(4):1289–1306, 2006.

Petros Drineas and Michael W. Mahoney. On the Nyström method for approximating a Gram matrix for improved kernel-based learning. *Journal of Machine Learning Research*, 6:2153–2175, 2005.

Rob Fergus, Yair Weiss, and Antonio Torralba. Semi-supervised learning in gigantic image collections. In *Neural Information Processing Systems*, 2009.

Charless Fowlkes, Serge Belongie, Fan Chung, and Jitendra Malik. Spectral grouping using the Nyström method. *Transactions on Pattern Analysis and Machine Intelligence*, 26(2):214–225, 2004.

A. Gustafson, E. Snitkin, S. Parker, C. DeLisi, and S. Kasif. Towards the identification of essential genes using targeted genome sequencing and comparative analysis. *BMC:Genomics*, 7:265, 2006.

Sanjiv Kumar, Mehryar Mohri, and Ameet Talwalkar. Ensemble Nyström method. In *Neural Information Processing Systems*, 2009.

Sanjiv Kumar, Mehryar Mohri, and Ameet Talwalkar. On sampling-based approximate spectral decomposition. In *International Conference on Machine Learning*, 2009.

Sanjiv Kumar, Mehryar Mohri, and Ameet Talwalkar. Sampling techniques for the Nyström method. In *Conference on Artificial Intelligence and Statistics*, 2009.

Yann LeCun and Corinna Cortes. The MNIST database of handwritten digits. `http://yann.lecun.com/exdb/mnist/`, 1998.

Andrew Y. Ng, Michael I. Jordan, and Yair Weiss. On spectral clustering: analysis and an algorithm. In *Neural Information Processing Systems*, pages 849–856, 2001.

John C. Platt. Fast embedding of sparse similarity graphs. In *Neural Information Processing Systems*, 2004.

Mark Rudelson. Random vectors in the isotropic position. *Journal of Functional Analysis*, 164(1):60–72, 1999.

Craig Saunders, Alexander Gammerman, and Volodya Vovk. Ridge Regression learning algorithm in dual variables. In *International Conference on Machine Learning*, 1998.

Bernhard Schölkopf, Alexander Smola, and Klaus-Robert Müller. Nonlinear component analysis as a kernel eigenvalue problem. *Neural Computation*, 10(5):1299–1319, 1998.

Terence Sim, Simon Baker, and Maan Bsat. The CMU pose, illumination, and expression database. In *Conference on Automatic Face and Gesture Recognition*, 2002.

Ameet Talwalkar, Sanjiv Kumar, and Henry Rowley. Large-scale manifold learning. In *Conference on Vision and Pattern Recognition*, 2008.

Christopher K. I. Williams and Matthias Seeger. Using the Nyström method to speed up kernel machines. In *Neural Information Processing Systems*, 2000.

Kai Zhang, Ivor Tsang, and James Kwok. Improved Nyström low-rank approximation and error analysis. In *International Conference on Machine Learning*, 2008.